# Minimum Enclosing Ball Synthetic Minority Oversampling Technique from a Geometric Perspective

Yi-Yang Shangguan, Shi-Shun Chen and Xiao-Yang Li

*Abstract*—Class imbalance refers to the significant difference in the number of samples from different classes within a dataset, making it challenging to identify minority class samples correctly. This issue is prevalent in real-world classification tasks, such as software defect prediction, medical diagnosis, and fraud detection. The synthetic minority oversampling technique (SMOTE) is widely used to address class imbalance issue, which is based on interpolation between randomly selected minority class samples and their neighbors. However, traditional SMOTE and most of its variants only interpolate between existing samples, which may be affected by noise samples in some cases and synthesize samples that lack diversity. To overcome these shortcomings, this paper proposes the Minimum Enclosing Ball SMOTE (MEB-SMOTE) method from a geometry perspective. Specifically, MEB is innovatively introduced into the oversampling method to construct a representative point. Then, high-quality samples are synthesized by interpolation between this representative point and the existing samples. The rationale behind constructing a representative point is discussed, demonstrating that the center of MEB is more suitable as the representative point. To exhibit the superiority of MEB-SMOTE, experiments are conducted on 15 real-world imbalanced datasets. The results indicate that MEB-SMOTE can effectively improve the classification performance on imbalanced datasets.

## I. INTRODUCTION

Class imbalance is the situation in a dataset used for classification tasks where the number of samples in one class significantly outnumbers the samples in other class. This can lead classifiers to be biased towards the majority class samples, making it challenging to accurately identify minority class samples [1]. Class imbalance issue is prevalent in real-world classification tasks, where the minority class samples typically represent critical events like defects [2], diseases [3], and fraud [4]. The failure to detect minority class samples can result in high error costs and risks [5]. Therefore, it is of great practical importance to improve the capability of classifier to identify minority class samples.

To address the class imbalance issue, existing research focuses on two levels: algorithmic-level [6] and data-level [7]. The algorithmic-level approaches aim to balance the impact of minority and majority class samples by adjusting class weights [8] or cost-sensitive learning [9], thereby improving the classification performance. However, algorithmic-level approaches depend on the setting of weights and costs and are usually specific to the classifier. In contrast, data-level approaches are independent of classifiers. Hence, the data-level approaches are more universally applicable. Data-level methods balance the number of samples from different classes through two strategies: undersampling and oversampling. Undersampling strategy balances the dataset by reducing the number of majority class samples, which unavoidably results in sample information loss [10]. On the other hand, oversampling strategy balances the dataset by increasing the number of minority class samples, which can maintain the rich sample information. Consequently, oversampling methods are widely used in real-world applications [11].

Among oversampling methods, the most straightforward method is random resampling, which increases the number of minority class samples in the training set by randomly duplicating the samples to achieve class balance. Obviously, it can easily result in overlapping samples, leading to overfitting issues [12]. To address this problem, Synthetic Minority Oversampling Technique (SMOTE) was proposed [13]. SMOTE is a classic oversampling method for class imbalance, which can effectively prevent the overfitting issues. It synthesizes new samples by randomly selecting a minority class sample and then performing linear interpolation between the selected sample and its nearest neighbors. This technique has been widely used and well-regarded in many fields [14].

To further improve the classification performance on imbalance datasets, several variant methods have been presented based on the traditional SMOTE. The most commonly used are Borderline-SMOTE [15] and ADASYN [16]. Borderline-SMOTE focuses on the boundary regions. By identifying borderline samples within the minority class, new samples are synthesized between these borderline samples and their neighbors. As for ADASYN, it sets a difficulty coefficient for each minority class sample to determine the number of new samples. Then, new samples are created by interpolation between the randomly selected sample and its neighbors. Both methods aim to improve the classifier's capacity by focusing on the samples that are difficult to classify during the training phase. In addition, some scholars improved the quality of synthesized samples by defining a geometric region and applying SMOTE within this region. For example, Georgios et al. [17] defined a hypersphere and used SMOTE to synthesize samples within this hypersphere region. Yuan et al. [18] constructed a convex hull for the set of minority class samples and identified a safe region by checking if the convex hull contains majority class samples. Then, new samples are generated within this region. These methods help reduce the occurrence of overlapping and noise samples to a certain extent.

While the above methods improve the performance of oversampling, they essentially still perform interpolation between two existing samples. When both existing samples are noise, the synthesized sample is also likely to be a noise sample, which will lead to the degradation of classification performance [19]. To avoid this issue, the interpolation rules between existing samples in the traditional SMOTE can be extended by replacing one of the existing samples with a new representative point, thus enhancing the robustness. For instance, the FCM-CSMOTE method proposed by Mohammed et al. [19] and the CP-SMOTE method proposed by Bao et al. [20] both selected the centroid point (the average position of all points in a set) as the representative point.

Although these centroid-based SMOTE methods construct a representative point for interpolation, they are easily affected by dense data regions, resulting in the synthesized samples concentrating around these regions. On the one hand, when there is a significant amount of noise samples surrounding the existing selected sample, the centroid point will inevitably be biased towards the dense noise region, leading to new noise samples. On the other hand, since the centroid point leans towards samples from dense regions, it may overlook minority class samples in sparse regions. This can lead to insufficient diversity of the synthesized samples, thereby affecting the generalization ability of the classifier.

To tackle the above challenges, this paper proposes a novel oversampling method based on the Minimum Enclosing Ball (MEB) and SMOTE from a geometric perspective, which is called MEB-SMOTE. MEB-SMOTE recursively calculates the MEB of a selected minority class sample and its neighbors. Then, the center of the MEB is used as the representative point and new samples are synthesized by interpolation between the selected sample and the center of the MEB. Since MEB-SMOTE captures the overall geometric structure of a sample set rather than simply relying on a single existing sample, it can enhance the robustness of the synthesized samples. Additionally, because dense regions are usually enclosed within the MEB, the center of the MEB is less affected by dense regions, improving the diversity of the synthesized samples. The contributions of this work are summarized as follows:

1) An improved SMOTE method is proposed by constructing a representative point for interpolation, which avoids creating new noise samples that result from direct interpolation between two existing noise samples.
2) The MEB is innovatively introduced for the first time to construct the representative point, which can better capture the geometric characteristic of the samples and improve the diversity of synthesized samples.
3) The proposed method is implemented in 15 real-world imbalanced datasets, showcasing its superiority over

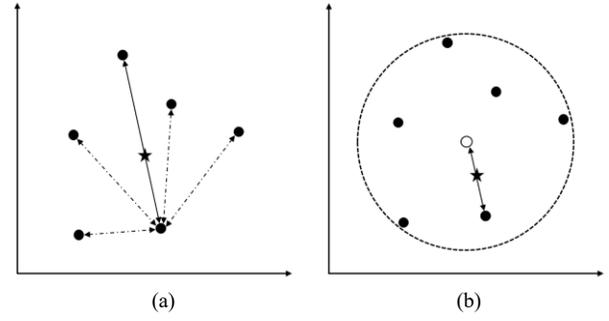

Fig. 1 Schematic depicting the new sample generation process: (a) SMOTE; (b) MEB-SMOTE.

traditional methods.

The rest of the paper is organized as follows. Section II introduces the necessary theoretical background, including the principles of SMOTE and MEB. The details of MEB-SMOTE are presented in Section III. In Section IV, MEB-SMOTE is implemented on several real-world imbalanced datasets, and the results of the experiments are discussed. Section V concludes this paper.

## II. PRELIMINARY

### A. Principles and Algorithm of the Traditional SMOTE

SMOTE proposed by Chawla et al. [13] achieves class balance in datasets by increasing the number of minority class samples. The essence of SMOTE is to synthesize new samples by interpolation between existing samples, instead of merely duplicating existing ones. Specifically, SMOTE selects the $k$-nearest neighbors of a minority class sample and then randomly interpolates between these samples to synthesize new ones, which can expand the minority class sample space and enable the classifier to learn more features of the minority class during the training phase.

Fig. 1 (a) illustrates the process of synthesizing a new sample using the traditional SMOTE. In Fig. 1 (a), the solid dots represent the selected existing sample and the neighbors, and the star dot represents a new synthesized sample. The detailed implementation steps of SMOTE are as follows.

Step 1: Calculate the number of samples to be synthesized and define the sample set. The imbalanced training dataset $\mathbf{X}$ contains both majority class samples and minority class samples. There are $N_{\text{maj}}$ majority class samples and $N_{\text{min}}$ minority class samples. The number of new minority class samples that need to be generated is $N_{\text{new}} = N_{\text{maj}} - N_{\text{min}}$. Hence, $N_{\text{new}}$ minority class samples can be randomly selected from $\mathbf{X}$ to form the set $\mathbf{X}_{\text{selected}}$.

Step 2: Determine the number of neighbors $k$ and compute the nearest neighbors. For each $x_i \in \mathbf{X}_{\text{selected}}$, compute its $k$-nearest neighbors (typically $k$=5) based on the Euclidean distance, forming the set $\mathbf{X}_i^k$. The Euclidean distance is calculated by:

$$d(x_i, x_j) = \sqrt{\sum_{p=1}^{n}(x_i^p - x_j^p)^2} \quad (1)$$

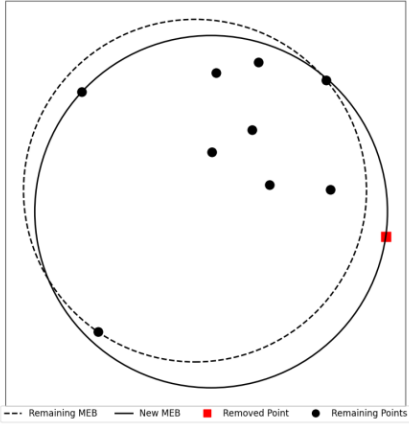

Fig. 2　Schematic using Welzl's algorithm to solve the MEB in 2-dimensional space.

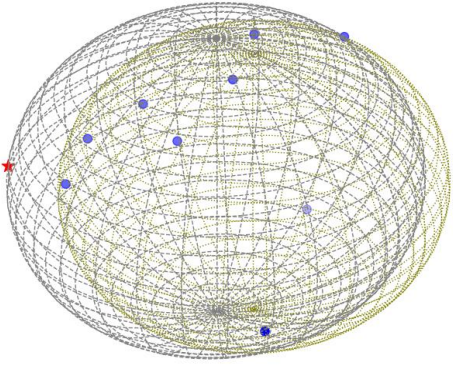

Fig. 3　Schematic using Welzl's algorithm to solve the MEB in 3-dimensional space.

where $d(x_i, x_j)$ represents the Euclidean distance between $x_i$ and $x_j$; and $n$ is the number of feature dimensions of the samples.

Step 3: Generate new samples through interpolation. Randomly select a sample $x_i^j$ from $\mathbf{X}_i^k$. Then, synthesize a new sample $x_{\text{new}}$ by linearly interpolating between $x_i$ and $x_i^j$. The formula for linear interpolation is:

$$x_{\text{new}} = x_i + \delta(x_i - x_i^j) \quad (2)$$

where $x_{\text{new}}$ represents a new synthesized sample; $\delta$ is a random number between 0 and 1.

Step 4: repeat step 2 and step 3 until the number of new synthesized samples $x_{\text{new}}$ reaches $N_{\text{new}}$, and form the set of new train samples $\mathbf{X}_{\text{new}}$.

### B. Principles and Algorithm of the MEB

MEB refers to a ball with the minimum radius which can enclose a given set of data. In high-dimensional space, it is a hypersphere. Specifically, the MEB can be described as an optimization problem: for a given set of data $S = \{x_1, x_2, ..., x_N\}$, the goal is to find a sphere $B(c, r)$ that satisfies the following condition:

$$\begin{aligned} \min \ & r \\ s.t. \ & \|x_i - c\| \leq r, \ x_i \in S \end{aligned} \quad (3)$$

where $c$ represents the center of MEB and $r$ represents the radius of MEB.

In order to solve the MEB, several methods have been developed in existing research. The most well-known method is Welzl's algorithm [21]. Welzl's algorithm is simple and efficient in implementation and can accurately compute the MEB of a finite point set in linear time. It performs exceptionally well on high-dimensional and large-scale datasets. Thus, Welzl's algorithm is used in this paper for solving MEB.

Specifically, Welzl's algorithm is an efficient method based on recursion and randomization for solving the MEB of a point set in an $n$-dimensional space. First, Welzl's algorithm defines some base cases. Then, by randomly selecting a point and removing it from the point set, it recursively computes the MEB of the remaining point set. If the removed point is within the MEB, this ball is the final MEB for all points. Otherwise, the point is added to the support set (the set of points currently assumed to be on the surface of the MEB) and the new point set is recursively computed. Fig. 2 and Fig. 3 illustrate the process of using Welzl's algorithm to solve the MEB for a set of samples in 2-dimensional and 3-dimensional spaces, respectively. In addition, the size of the support set cannot exceed $n+1$. Because at most $n+1$ points determine a unique MEB in $n$-dimensional space. The specific implementation details of Welzl's method are available in [21].

In fact, MEB has been applied in various fields, such as cluster analysis [22] and anomaly detection [23]. However, no research has applied it yet to address the class imbalance issue. This paper is the first to introduce the concept and the algorithm of MEB into the SMOTE method. To enhance the quality of synthesized samples, the center of MEB is used as a representative point for a sample set, with Welzl's algorithm solving the MEB.

## III. PROPOSED METHOD

### A. MEB-SMOTE

The critical idea of the MEB-SMOTE is to select the MEB center as the representative point and interpolate with this representative point to synthesize samples, thus expanding the rules of the traditional SMOTE.

First, MEB-SMOTE randomly selects a minority class sample. Next, the nearest neighbors of the selected sample are calculated and forms a set. To better represent the characteristics and geometric structure of the set, the MEB center is used as its representative point, and Welzl's algorithm is used to calculate the MEB of the set. Subsequently, the new samples are generated through linear interpolation between the existing selected sample and the representative point. The detailed steps are as follows.

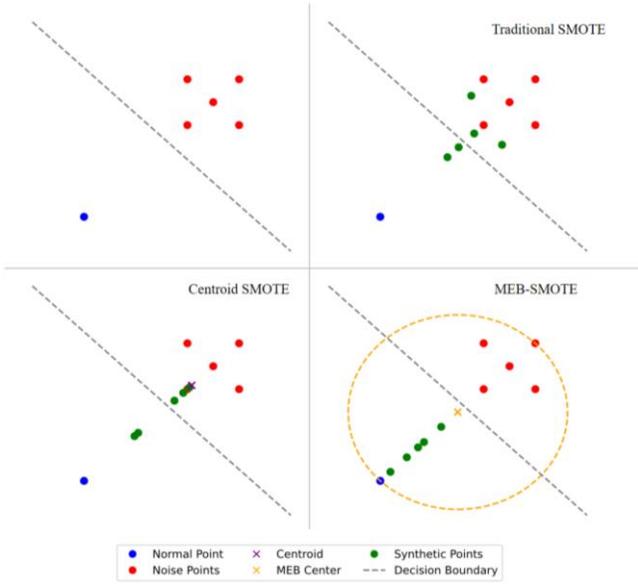

Fig. 4 Comparison of samples synthesized by MEB-SMOTE with samples synthesized by traditional SMOTE and centroid SMOTE methods in two-dimensional space.

---

Algorithm 1: MEB-SMOTE ($N_{maj}$, $N_{min}$, **X**, $k$)

**Input:** number of majority class samples $N_{maj}$; number of minority class samples $N_{min}$; imbalanced training dataset **X**; number of nearest neighbors to consider $k$
**Output:** a set of new generated samples $\mathbf{X}_{new}^{MEB}$
**Begin Pseudo-Code**
1. Calculate $N_{new} = N_{maj} - N_{min}$
2. Randomly select $N_{new}$ minority class samples from **X** to form the set $\mathbf{X}_{selected}$
3. **for** each $x_i \in \mathbf{X}_{selected}$ do
   4. Compute the $k$ nearest neighbors of $x_i$ to form the set $\mathbf{X}_i^k$ (usually $k=5$)
   5. Solve the $MEB(c_i, r_i)$ of $\mathbf{X}_i^k$ by Welzl's algorithm
   6. Generate a random number $\alpha$ between 0 and 1
   7. Synthesize a sample $x_{new}^{MEB}$ through linear interpolation between $x_i$ and $c_i$ with the random number $\alpha$
   8. Add the new sample $x_{new}^{MEB}$ to the set $\mathbf{X}_{new}^{MEB}$
9. **End for**
**End of Pseudo-Code**

---

Step 1: Construct the sample set to be processed. First, the imbalanced training dataset **X** contains $N_{maj}$ majority class samples and $N_{min}$ minority class samples. The number of new samples that need to be generated is $N_{new} = N_{maj} - N_{min}$. Then, randomly select $N_{new}$ minority class samples from **X** to form the set $\mathbf{X}_{selected}$. For each $x_i \in \mathbf{X}_{selected}$, the $k$ nearest neighbors can be calculated based on Euclidean distance, forming the set $\mathbf{X}_i^k$. The Euclidean distance can be calculated by (1).

Step 2: The MEB of $\mathbf{X}_i^k$ can be solved by Welzl's algorithm as described in Section II, and it is denoted as $MEB(c_i, r_i)$. Where, $c_i$ is the center of the MEB and $c_i$ is used as the representative point of point set $\mathbf{X}_i^k$, $r_i$ is the radius of the MEB.

Step 3: Synthesize new samples through interpolation. Perform linear interpolation between $c_i$ solved by Step 2 and the selected existing sample $x_i$ to synthesize a new sample $x_{new}^{MEB}$. The formula for linear interpolation is:

$$x_{new}^{MEB} = x_i + \alpha(x_i - c_i) \tag{4}$$

where $x_{new}^{MEB}$ represents a new sample synthesized by the MEB-SMOTE; $\alpha$ is a random number between 0 and 1.

Finally, repeat the step 2 and step 3 until the number of synthesized samples $x_{new}^{MEB}$ reaches $N_{new}$. The new generated samples form the set $\mathbf{X}_{new}^{MEB}$, which can be used as the new train dataset.

Fig. 1 (b) shows the process of MEB-SMOTE synthesizing a new sample. In Fig. 1 (b), the solid dots represent the selected sample and the neighbors, the hollow dot represents the center of the MEB and the star dot represents a new synthesized sample.

The pseudo-code of MEB-SMOTE is shown in Algorithm 1.

### B. Rationality Analysis

As mentioned above, the traditional SMOTE and some of its variants interpolate between two randomly selected existing minority class samples. Sometimes this can result in noise samples. Additionally, replacing one of the existing samples with a representative point can improve the synthesized samples to some extent. However, using only the centroid as the representative point and interpolating between it and the selected existing sample can be affected by dense regions. Conversely, MEB-SMOTE uses the center of MEB as the representative point, interpolating between this MEB center and the selected existing sample to synthesize both robust and diverse samples.

To compare the quality of samples generated by traditional SMOTE, centroid SMOTE, and MEB-SMOTE, the following scenario is considered in this paper.

First, assume that the selected minority class samples and their neighbors are the same for different oversampling methods. The selected samples are correctly labeled (normal samples), but their neighbors are incorrectly labeled (noise samples), which can simulate the situation when the quality of the existing data is poor. Under this circumstance, traditional SMOTE and centroid SMOTE strategies are biased to synthesize more noise samples because of the impact of the dense noise region. In contrast, MEB-SMOTE is more likely to generate the normal samples.

Fig. 4 shows the samples generated by different oversampling strategies in this case. Where, the blue dots represent normal samples; the red dots represent noise samples, which are the neighbors of the blue dots. The purple dot represents the centroid of the blue and red dots. The yellow dot represents the center of the MEB, and the green dots represent the new synthesized samples. The gray dashed line represents the classification boundary.

TABLE I
DETAILED INFORMATION OF THE DATASET

| Name | F&S | Att | Min | Maj | IR |
|---|---|---|---|---|---|
| jedit-3.2 | SDP (PROMISE) | 20 | 90 | 182 | 2.02 |
| jedit-4.0 | SDP (PROMISE) | 20 | 75 | 231 | 3.08 |
| ant-1.7 | SDP (PROMISE) | 20 | 166 | 579 | 3.49 |
| ivy-2.0 | SDP (PROMISE) | 20 | 40 | 312 | 7.80 |
| poi-2.0 | SDP (PROMISE) | 20 | 37 | 277 | 7.49 |
| camel-1.6 | SDP (PROMISE) | 20 | 188 | 777 | 4.13 |
| velocity-1.6 | SDP (PROMISE) | 20 | 78 | 151 | 1.94 |
| KC1 | SDP (NASA) | 21 | 294 | 868 | 2.95 |
| PC1 | SDP (NASA) | 37 | 55 | 624 | 11.35 |
| bupa-liver | MD (UCI) | 6 | 145 | 200 | 1.38 |
| wdbc | MD UCI | 30 | 212 | 357 | 1.68 |
| creditcard | FFD (UCI) | 30 | 492 | 284315 | 577.88 |
| manufacturing | IFD (Kaggle) | 16 | 517 | 2723 | 5.27 |
| ionosphere | other (UCI) | 34 | 126 | 225 | 1.79 |
| pageblocks | other (UCI) | 10 | 560 | 4913 | 8.77 |

F&S stands for field and source, ATT is the attribute of the dataset, min is the number of minority class samples, maj is the number of majority class samples.

From Fig. 4, it can be deduced that:

1. In the traditional SMOTE, the synthesized samples are mainly distributed near the noise samples. This indicates that when noise samples are near the selected samples, the traditional SMOTE method is likely to interpolate with the noise samples, which may potentially introduce more noise samples and thus reducing the data quality.

2. The strategy of the centroid SMOTE is to select the centroid of a set as the representative point. Specifically, the centroid of the K samples $P = \{p_1, p_2, ..., p_K\}$ can be calculated by:

$$centroid(P) = \frac{1}{K}\sum_{i=1}^{K} p_i \quad (5)$$

where centroid(P) denotes the centroid point of the set P.

After that, use this centroid point to interpolate with the selected sample, thereby generating new samples. Although the centroid SMOTE strategy selects a representative point, it avoids direct interpolation with noise samples to some extent. As shown in (5), the centroid is easily influenced by majority data. In such a case, the distribution of the centroid will be more biased towards the noise region. Consequently, if the selected sample is interpolated with the centroid point, the generated samples may also be near the noise region.

3. As for the MEB-SMOTE method, even when there are many noise samples among the neighbors, the center of the MEB is not biased towards dense noise regions. This is because the center of the MEB is usually determined by boundary points and is hardly influenced by internal points. As a result, the distribution of the MEB center can be close to the normal region. Therefore, interpolating with the MEB center synthesizes more normal samples and reduces the influence of dense noise samples.

Furthermore, the dense distribution of neighbors in Fig. 4 can still cause problems, even if they are not noisy samples. Traditional SMOTE and centroid SMOTE are influenced by dense regions, making the synthesized samples more likely to be close to these dense regions. This issue can lead to an over-concentration effect. More synthesized samples appear in dense regions, while sparse regions generate few new samples This over-concentration effect indicates insufficient diversity and ineffective coverage of the sample distribution. In contrast, MEB-SMOTE can synthesize samples that better fill the sparse regions, balancing the existing sample distribution. It can result in more diverse samples and enhance the generalization capability of the classifiers.

IV. EXPERIMENTS

A. Evaluation Metrics

In traditional binary classification tasks, the minority class samples are usually labeled as positive and the majority class samples are labeled as negative [24]. Additionally, the confusion matrix is used to evaluate classification performance, which consists of TP, FP, FN, and TN. TP represents the number of correctly predicted positive samples; FP represents the number of samples incorrectly predicted as positive; FN represents the number of samples incorrectly predicted as negative; and TN represents the number of correctly predicted negative samples.

The well-known accuracy (ACC) is the most commonly used evaluation metric, as shown in (6).

$$ACC = \frac{TP+TN}{TP+TN+FP+FN} \quad (6)$$

However, in some imbalanced datasets, accuracy is easily dominated by majority class samples, making it challenging to reflect the classification performance comprehensively.

Besides ACC, the F1 score (F1) is also a commonly used performance evaluation metric for classification tasks, particularly suitable for imbalanced datasets. It can comprehensively consider precision and recall. Specifically, precision can be calculated by (7). Recall can be calculated by (8). F1 is the harmonic mean of precision and recall and is calculated by (9).

$$Precision = \frac{TP}{TP+FP} \quad (7)$$

$$Recall = \frac{TP}{TP+FN} \quad (8)$$

$$F1 = 2 \times \frac{Precision \times Recall}{Precision + Recall} \quad (9)$$

Furthermore, the AUC (Area Under the Curve) is crucial for imbalanced classification and is one of the most significant metrics for evaluating imbalanced classification performance comprehensively [25]. It represents the area under the ROC (Receiver Operating Characteristic) curve, which plots the true positive rate (TPR) against the false positive rate (FPR) at various threshold settings.

Therefore, ACC, F1, and AUC are selected as the evaluation metrics for the experiments, to comprehensively and objectively reflect classification performance.

## B. Datasets

In the experiments, the proposed MEB-SMOTE method is tested on 15 real-world datasets, encompassing a diverse range of fields, including software defect prediction (SDP), medical diagnosis (MD), financial fraud detection (FFD), industrial fault detection (IFD), and others. The datasets used in the experiments are sourced from well-known repositories such as PROMISE, NASA, UCI, and Kaggle, which ensures a comprehensive evaluation of the performance across different domains and types of data.

These datasets all have class imbalance issues with different imbalance ratio (IR). The degree of imbalance is usually represented by the IR [26], which is the size of the majority class divided by the size of the minority class. The higher the IR, the greater the degree of imbalance. These datasets cover a wide range of application situations, with IR ranging from 1.38 to 577.88 to thoroughly validate the effectiveness of the proposed method. Detailed information about the datasets can be found in Table I.

## C. Experimental Setup

To validate the effectiveness of the MEB-SMOTE proposed in this paper, a basic classifier needs to be selected to classify the samples. According to [27], XGBoost (Extreme Gradient Boosting) is deemed suitable as the basic classifier for the oversampling method, which is widely used in various classification tasks [28].

Furthermore, the following methods are used as the comparisons in experiments. First, the traditional SMOTE method can be used as a baseline to evaluate the innovations and improvements of the proposed MEB-SMOTE. Second, since Borderline-SMOTE and ADASYN are the most widely applied variants of SMOTE in current research, MEB-SMOTE is also compared with them. Additionally, to comprehensively demonstrate the superiority of MEB-SMOTE, it will be compared against the centroid SMOTE method. Specifically, the traditional SMOTE, Borderline-SMOTE and ADASYN are implemented using the existing imbalanced-learn toolbox [29]. The centroid SMOTE method is similar to the MEB-SMOTE process proposed in this paper, with the centroid chosen as the representative point. In centroid SMOTE, the $k$-nearest neighbors of the selected minority class sample are computed, and then the centroid of this set of data can be calculated. New samples are synthesized by interpolating between the selected minority class samples and the centroid. As for the MEB-SMOTE method, the process is implemented as described in Algorithm 1. These comparisons can help demonstrate the superiority of MEB-SMOTE.

The number of neighbors for all methods is set to $k$=5. Then, the four oversampling methods are applied to the datasets separately to generate new training sets of minority class. Next, the XGBoost classifier is trained on the new sets processed by the different oversampling methods. After that, the classification performance in the test set is evaluated.

TABLE II
ACC RESULTS OF DIFFERENT DATASET WITH DIFFERENT OVERSAMPLING METHODS

| Name | ACC | | | | F1 | | | | AUC | | | |
|---|---|---|---|---|---|---|---|---|---|---|---|---|
| | SMOTE | ADASYN | Centroid SMOTE | MEB-SMOTE | SMOTE | ADASYN | Centroid SMOTE | MEB-SMOTE | SMOTE | ADASYN | Centroid SMOTE | MEB-SMOTE |
| jedit-3.2 | 0.7681 ±0.04 | 0.7681 ±0.03 | 0.7756 ±0.03 | **0.7832** ±0.03 | 0.6518 ±0.05 | 0.6541 ±0.05 | 0.6519 ±0.04 | **0.6644** ±0.06 | 0.8376 ±0.03 | 0.8141 ±0.03 | 0.8295 ±0.04 | **0.8412** ±0.04 |
| jedit-4.0 | 0.7745 ±0.03 | 0.7580 ±0.04 | 0.8071 ±0.04 | **0.8463** ±0.02 | 0.5686 ±0.09 | 0.5295 ±0.07 | 0.5976 ±0.08 | **0.6560** ±0.05 | 0.8179 ±0.07 | 0.8228 ±0.06 | 0.8360 ±0.0 | **0.8457** ±0.06 |
| ant-1.7 | 0.8054 ±0.02 | 0.8067 ±0.02 | 0.7987 ±0.03 | **0.8188** ±0.03 | 0.5773 ±0.02 | 0.5773 ±0.03 | 0.5342 ±0.07 | **0.5803** ±0.05 | 0.8143 ±0.02 | 0.8135 ±0.03 | 0.8157 ±0.02 | **0.8206** ±0.03 |
| ivy-2.0 | 0.8579 ±0.03 | 0.8381 ±0.04 | 0.8665 ±0.04 | **0.8778** ±0.04 | **0.4271** ±0.07 | 0.3789 ±0.10 | 0.3029 ±0.14 | 0.3929 ±0.12 | 0.7697 ±0.09 | 0.7591 ±0.10 | 0.7591 ±0.10 | **0.7859** ±0.10 |
| poi-2.0 | 0.8311 ±0.01 | 0.8311 ±0.02 | 0.8630 ±0.02 | **0.8599** ±0.04 | 0.2808 ±0.10 | 0.3071 ±0.09 | 0.3154 ±0.09 | **0.3607** ±0.12 | 0.7349 ±0.08 | 0.7425 ±0.08 | 0.7533 ±0.09 | **0.7570** ±0.09 |
| camel-1.6 | 0.7627 ±0.02 | 0.7617 ±0.02 | 0.7845 ±0.02 | **0.7907** ±0.01 | 0.3506 ±0.06 | 0.3471 ±0.06 | 0.3290 ±0.08 | **0.3660** ±0.05 | 0.7209 ±0.06 | 0.7105 ±0.04 | 0.7181 ±0.03 | **0.7300** ±0.04 |
| velocity-1.6 | 0.7507 ±0.06 | 0.7376 ±0.06 | 0.7592 ±0.08 | **0.7637** ±0.07 | 0.6100 ±0.11 | 0.5966 ±0.11 | 0.6122 ±0.14 | **0.6428** ±0.11 | 0.8302 ±0.07 | 0.8111 ±0.08 | 0.8201 ±0.07 | **0.8317** ±0.06 |
| KC1 | 0.7298 ±0.02 | 0.7246 ±0.01 | 0.7599 ±0.01 | **0.7694** ±0.01 | 0.4678 ±0.03 | 0.4565 ±0.03 | 0.4561 ±0.05 | **0.4972** ±0.04 | 0.7190 ±0.02 | 0.7086 ±0.01 | 0.7212 ±0.02 | **0.7257** ±0.02 |
| PC1 | 0.8984 ±0.02 | 0.8999 ±0.03 | **0.9205** ±0.01 | 0.9116 ±0.02 | 0.3610 ±0.06 | 0.3817 ±0.06 | 0.3798 ±0.11 | **0.4073** ±0.13 | 0.8426 ±0.03 | 0.8475 ±0.02 | 0.8683 ±0.03 | **0.8727** ±0.02 |
| bupa-liver | 0.6957 ±0.08 | 0.6812 ±0.07 | 0.7072 ±0.10 | **0.7217** ±0.07 | 0.6250 ±0.10 | 0.6227 ±0.08 | 0.6548 ±0.09 | **0.6591** ±0.07 | 0.7590 ±0.09 | 0.7567 ±0.08 | 0.7650 ±0.08 | **0.7671** ±0.07 |
| wdbc | 0.9578 ±0.02 | 0.9614 ±0.02 | 0.9614 ±0.01 | **0.9701** ±0.00 | 0.9440 ±0.02 | 0.9490 ±0.02 | 0.9487 ±0.01 | **0.9595** ±0.01 | 0.9920 ±0.01 | 0.9919 ±0.01 | 0.9931 ±0.00 | **0.9941** ±0.00 |
| creditcard | **0.9995** ±0.00 | **0.9995** ±0.00 | **0.9995** ±0.00 | **0.9995** ±0.00 | 0.8485 ±0.02 | 0.8483 ±0.02 | 0.8627 ±0.02 | **0.8571** ±0.03 | 0.9811 ±0.00 | 0.9774 ±0.01 | 0.9773 ±0.01 | **0.9883** ±0.00 |
| manufacturing | 0.9448 ±0.01 | 0.9429 ±0.01 | 0.9586 ±0.01 | **0.9600** ±0.01 | 0.8183 ±0.04 | 0.8131 ±0.03 | 0.8565 ±0.04 | **0.8587** ±0.04 | 0.8901 ±0.03 | 0.8918 ±0.03 | 0.8937 ±0.03 | **0.8942** ±0.02 |
| ionosphere | 0.9245 ±0.04 | 0.9116 ±0.02 | 0.9373 ±0.03 | **0.9434** ±0.03 | 0.8919 ±0.05 | 0.8751 ±0.03 | 0.9096 ±0.04 | **0.9211** ±0.04 | 0.9644 ±0.02 | 0.9606 ±0.02 | 0.9735 ±0.02 | **0.9812** ±0.02 |
| pageblocks | 0.9675 ±0.01 | 0.9666 ±0.01 | **0.9720** ±0.00 | 0.9715 ±0.00 | 0.8446 ±0.03 | 0.8420 ±0.03 | **0.8616** ±0.02 | 0.8596 ±0.01 | 0.9852 ±0.01 | 0.9851 ±0.00 | 0.9867 ±0.00 | **0.9877** ±0.01 |

To ensure the reliability of the experimental results, 5-fold cross-validation is used to assess the model performance. Finally, the mean and standard deviation of the ACC, F1, and AUC metrics are calculated to measure the effectiveness of different oversampling methods.

*D. Results*

This section presents the detailed experimental results on 15 different datasets. Specifically, the mean and standard deviation of the evaluation metrics for each oversampling method on each dataset are calculated, providing a more comprehensive and detailed comparison. To highlight the best-performing oversampling method, the same classifier is used for each dataset, and the best evaluation metric result among the four oversampling methods is marked in bold, as shown in Table II.

From the results above, it can be seen that the MEB-SMOTE method demonstrated superior classification performance on most datasets. In 13 datasets, the MEB-SMOTE method achieved the highest ACC and F1 scores, and achieved the highest AUC in all 15 datasets

Furthermore, in some cases, the centroid SMOTE performs better than the traditional SMOTE and ADASYN. This suggests that the strategy of selecting representative points for interpolation is effective, which may generate higher-quality samples. However, the performance of the centroid SMOTE method is still inferior to the MEB-SMOTE proposed in this paper. It indicates that the center point of MEB is more suitable as the representative point than the centroid point. This is because the centroid point assumes that each data point has equal importance, making it susceptible to adverse effects from dense regions. In contrast, the MEB center point pays more attention to the overall geometric structure of the data and has a more global perspective, which can prevent the synthesized samples from being overly concentrated in the existing dense regions.

Additionally, the standard deviation of the evaluation metrics shows that the MEB-SMOTE method does not increase the standard deviation compared to traditional methods. It indicates that the stability of the four methods is similar. This is also particularly important because the standard deviation can reflect the consistency and reliability of the methods across different experiments. A smaller standard deviation means that the method performs more stably and predictably across different datasets.

In summary, the MEB-SMOTE method demonstrates significant advantages in synthesizing new samples. It can synthesize more robust and diverse minority class samples, which helps the classifier more effectively learn the characteristics of different classes, thereby improving overall performance. Meanwhile, the experimental results show that MEB-SMOTE exhibits significant advantages in handling imbalanced classification problems, not only outperforming traditional methods in multiple evaluation metrics but also maintaining high stability.

## V. CONCLUSION

Class imbalance is prevalent in the real world, making it challenging to identify minority class samples correctly. SMOTE and some of its variants offer solutions for this issue. However, these methods may generate noise samples in some cases and result in insufficient sample diversity. To address these limitations and improve classification performance, this paper proposes a novel oversampling method called MEB-SMOTE.

In the MEB-SMOTE method, the first step is to construct the set of minority class samples to be processed. Then, the MEB of this set can be solved by Welzl's algorithm. Next, new minority class samples can be synthesized through interpolation between the existing minority class sample and the MEB center. Furthermore, the rationale for constructing the MEB center as the representative point is discussed. It can be inferred that the characteristics of the minority class sample set can be better captured by the MEB center from a geometric perspective, and it is less influenced by the dense samples. Consequently, the MEB center is considered a more suitable representative point for interpolation. Finally, the effectiveness of MEB-SMOTE is evaluated on 15 real-world datasets. The results demonstrate that MEB-SMOTE outperforms the baseline methods and centroid SMOTE in most cases, showcasing the superiority of MEB-SMOTE.

Additionally, although centroid SMOTE performs worse than the proposed MEB-SMOTE, it still outperforms traditional methods in some cases. This indicates that selecting a representative point for interpolation is an effective strategy, rather than merely interpolating between existing samples. This insight suggests a promising direction for future research: developing more effective representative points to further enhance oversampling methods for class imbalance. This also suggests that developing the more effective representative point could be a possible direction for further research, improving oversampling methods on imbalanced datasets.

REFERENCES

[1] H. He, and E. A. Garcia, "Learning from Imbalanced Data," *IEEE Trans. Knowl. Data Eng.,* vol. 21, no. 9, pp. 1263-1284, Sep. 2009.
[2] Q. Song, Y. Guo, and M. Shepperd, "A Comprehensive Investigation of the Role of Imbalanced Learning for Software Defect Prediction," *IEEE Trans. Softw. Eng.,* vol. 45, no. 12, pp. 1253-1269, Dec. 2019.
[3] Z. Xu, D. Shen, T. Nie, Y. Kou, N. Yin, and X. Han, "A cluster-based oversampling algorithm combining SMOTE and k-means for imbalanced medical data," *Inf. Sci.,* vol. 572, pp. 574-589, Sep. 2021.
[4] S. Makki, Z. Assaghir, Y. Taher, R. Haque, M. S. Hacid, and H. Zeineddine, "An Experimental Study With Imbalanced Classification Approaches for Credit Card Fraud Detection," *IEEE Access,* vol. 7, pp. 93010-93022, Jul. 2019.
[5] G. Haixiang, L. Yijing, J. Shang, G. Mingyun, H. Yuanyue, and G. Bing, "Learning from class-imbalanced data: Review of methods and applications," *Expert Syst. Appl.,* vol. 73, pp. 220-239, May. 2017.
[6] B. Krawczyk, "Learning from imbalanced data: open challenges and future directions," *Prog. Artif. Intell.,* vol. 5, no. 4, pp. 221-232, Nov. 2016.
[7] J. M. Johnson, and T. M. Khoshgoftaar, "Survey on deep learning with class imbalance," *J. Big Data.,* vol. 6, no. 1, pp. 27, Mar. 2019.
[8] K. R. M. Fernando, and C. P. Tsokos, "Dynamically Weighted Balanced Loss: Class Imbalanced Learning and Confidence Calibration of Deep Neural Networks," *IEEE Trans. Neural Netw. Learn. Syst.,* vol. 33, no. 7, pp. 2940-2951, Jul. 2022.
[9] Z. Zhi-Hua, and L. Xu-Ying, "Training cost-sensitive neural networks with methods addressing the class imbalance problem," *IEEE Trans. Knowl. Data Eng.,* vol. 18, no. 1, pp. 63-77, Jan. 2006.
[10] T. Hasanin, and T. Khoshgoftaar, "The Effects of Random Undersampling with Simulated Class Imbalance for Big Data," in *2018 IEEE Int. Conf. Inf. Reuse Integr. (IRI)*, 2018, pp. 70-79.

[11] R. Mohammed, J. Rawashdeh, and M. Abdullah, "Machine Learning with Oversampling and Undersampling Techniques: Overview Study and Experimental Results," in *2020 11th Int. Conf. Inf. Commun. Syst. (ICICS)*, 2020, pp. 243-248.

[12] Z. Wang, T. Liu, X. Wu, and C. Liu, "Application of an Oversampling Method Based on GMM and Boundary Optimization in Imbalance-Bearing Fault Diagnosis," *IEEE Trans. Ind. Informat.,* vol. 20, no. 2, pp. 1931-1940, Feb. 2024.

[13] N. V. Chawla, K. W. Bowyer, L. O. Hall, and W. P. Kegelmeyer, "SMOTE: synthetic minority over-sampling technique," *J. Artif. Intell. Res.,* vol. 16, pp. 321-357, Jun. 2002.

[14] J. Li, Q. Zhu, Q. Wu, and Z. Fan, "A novel oversampling technique for class-imbalanced learning based on SMOTE and natural neighbors," *Inf. Sci.,* vol. 565, pp. 438-455, Jul. 2021.

[15] H. Han, W.-Y. Wang, and B.-H. Mao, "Borderline-SMOTE: A New Over-Sampling Method in Imbalanced Data Sets Learning," in *Adv. Intell. Comput.*, Berlin, Heidelberg, 2005, pp. 878-887.

[16] H. Haibo, B. Yang, E. A. Garcia, and L. Shutao, "ADASYN: Adaptive synthetic sampling approach for imbalanced learning," in *2008 IEEE Int. Jt. Conf. Neural Netw. (IEEE World Congr. Comput. Intell.)*, 2008, pp. 1322-1328.

[17] G. Douzas, and F. Bacao, "Geometric SMOTE a geometrically enhanced drop-in replacement for SMOTE," *Inf. Sci.,* vol. 501, pp. 118-135, Oct. 2019.

[18] X. Yuan, S. Chen, H. Zhou, C. Sun, and L. Yuwen, "CHSMOTE: Convex hull-based synthetic minority oversampling technique for alleviating the class imbalance problem," *Inf. Sci.,* vol. 623, pp. 324-341, Apr. 2023.

[19] R. Mohammed, and E. M. Karim, "FCM-CSMOTE: Fuzzy C-Means Center-SMOTE," *Expert Syst. Appl.,* vol. 248, pp. 123406, Aug. 2024.

[20] Y. Bao, and S. Yang, "Two Novel SMOTE Methods for Solving Imbalanced Classification Problems," *IEEE Access,* vol. 11, pp. 5816-5823, Jan. 2023.

[21] E. Welzl, "Smallest enclosing disks (balls and ellipsoids)," in *New Results New Trends Comput. Sci.*, Berlin, Heidelberg, 1991, pp. 359-370.

[22] P. Qian, F. L. Chung, S. Wang, and Z. Deng, "Fast Graph-Based Relaxed Clustering for Large Data Sets Using Minimal Enclosing Ball," *IEEE Trans. Syst. Man Cybern.,* vol. 42, no. 3, pp. 672-687, Feb. 2012.

[23] L. Mou, Y. Hua, S. Saha, F. Bovolo, L. Bruzzone, and X. X. Zhu, "Detecting Changes by Learning No Changes: Data-Enclosing-Ball Minimizing Autoencoders for One-Class Change Detection in Multispectral Imagery," *IEEE Trans. Geosci. Remote Sens.,* vol. 60, pp. 1-16, Aug. 2022.

[24] L. Zhu, C. Lu, Z. Y. Dong, and C. Hong, "Imbalance Learning Machine-Based Power System Short-Term Voltage Stability Assessment," *IEEE Trans. Ind. Informat.,* vol. 13, no. 5, pp. 2533-2543, Apr. 2017.

[25] J. H. Xue, and P. Hall, "Why Does Rebalancing Class-Unbalanced Data Improve AUC for Linear Discriminant Analysis?," *IEEE Trans. Pattern Anal. Mach. Intell.,* vol. 37, no. 5, pp. 1109-1112, Sep. 2015.

[26] M. Krstic, and M. Bjelica, "Impact of class imbalance on personalized program guide performance," *IEEE Trans. Consum. Electron.,* vol. 61, no. 1, pp. 90-95, Feb. 2015.

[27] Z. Wu, W. Lin, B. Fu, J. Guo, Y. Ji, and M. Pecht, "A Local Adaptive Minority Selection and Oversampling Method for Class-Imbalanced Fault Diagnostics in Industrial Systems," *IEEE Trans. Reliab.,* vol. 69, no. 4, pp. 1195-1206, Dec. 2020.

[28] T. Chen, and C. Guestrin, "Xgboost: A scalable tree boosting system," in *Proc. 22nd ACM SIGKDD Int. Conf. Knowl. Discov. Data Min.*, 2016, pp. 785-794.

[29] G. LemaÃŽtre, F. Nogueira, and C. K. Aridas, "Imbalanced-learn: A python toolbox to tackle the curse of imbalanced datasets in machine learning," *Journal of machine learning research,* vol. 18, no. 17, pp. 1-5. 2017.